\documentclass{llncs}
\usepackage{amsmath, graphicx, booktabs, tabularx}
\usepackage{hyperref}
\usepackage{wrapfig}

\graphicspath{{./figures/}}
\begin{document}

\title{Bimodal network architectures for automatic generation of image annotation from text}
\author{Mehdi Moradi\thanks{Corresponding author}, Ali Madani, Yaniv Gur, Yufan Guo, Tanveer Syeda-Mahmood}
\institute{IBM Research - Almaden Research Center \\ \email{mmoradi@us.ibm.com}}

\maketitle              

\begin{abstract}

Medical image analysis practitioners have embraced big data methodologies. This has created a need for large annotated datasets. The source of big data is typically large image collections and clinical reports recorded for these images. In many cases, however, building algorithms aimed at segmentation and detection of disease requires a training dataset with markings of the areas of interest on the image that match with the described anomalies. This process of annotation is expensive and needs the involvement of clinicians. In this work we propose two separate deep neural network architectures for automatic marking of a region of interest (ROI) on the image best representing a finding location, given a textual report or a set of keywords. One architecture consists of LSTM and CNN components and is trained end to end with images, matching text, and markings of ROIs for those images. The output layer estimates the coordinates of the vertices of a polygonal region. The second architecture uses a network pre-trained on a large dataset of the same image types for learning feature representations of the findings of interest. We show that for a variety of findings from chest X-ray images, both proposed architectures learn to estimate the ROI, as validated by clinical annotations. There is a clear advantage obtained from the architecture with pre-trained imaging network. The centroids of the ROIs marked by this network were on average at a distance equivalent to 5.1\% of the image width from the centroids of the ground truth ROIs. 

\end{abstract}
\section{Introduction}

Big data methods such as deep learning consume large quantities of labeled data. In medical imaging, such big data approaches have shown great impact in advancing segmentation and disease detection algorithms. To expedite the progress of this area of research, access to such datasets is vital. Much of the radiology data is recorded as paired image and radiology reports. While these are useful for training purposes and can be mined for labels, image level annotations are rarely recorded in clinical practice. Therefore, even though the diagnosis or finding is already described by a clinician in a report, clinical input is still required in marking the area depicting the finding back onto the image. In some cases, such as mass or nodules of the lung in chest X-ray images, the finding could be limited to a small portion of a large organ. 

In this work we propose to estimate the position of the region of interest (ROI) described in a text segment, written for a medical image, using a supervised neural network architecture. This area of work is fresh in medical image analysis. Attention networks have been proposed for highlighting an ROI in an image that contributes most to the output of a convolutional neural network (CNN). These have shown impressive outcomes in applications such as image captioning \cite{DBLP:journals/corr/XuBKCCSZB15}. However, attention methods assume the availability of a detector and use only the image as the input, whereas our contribution is in facilitating the data curation step given the text reports. Other related work in data curation for medical image analysis include image to text mapping for automatic label generation for images \cite{moradi2016cross}. This method does not mark ROIs on the image. There is also some work in crowd-sourcing of medical image annotations \cite{Rodríguez2012,LenaMaierHein2014}. The automatic approach in the current paper can be used as a complement to crowd-sourcing where images can be annotated with some preliminary contours before being edited by human annotators.

The main contributions of this paper are bimodal (text+image) neural network architectures that estimate the ROIs coordinates, given an image and a text segment. We propose two bimodal architectures to facilitate two different common scenarios. \textbf{I)} We train an end to end network that can process text using recurrent layers, and the images using convolutional layers, and merge the two to estimate the ROI coordinates. This integrated network is, however, difficult to train with limited annotated data. It is particularly difficult to sufficiently characterize the visual signature of relevant findings using limited data. \textbf{II)} Since the main goal here is to limit the effort to annotate large quantities of images, we propose an alternative for situations when an organized collection of images, text and markings is scarce. This second approach separates the training of the image processing CNN, and the text processing network, from the network that maps the concatenated image and text features to the ROI coordinate space. The imaging network is trained on a very large public dataset to learn the features characterizing clinical findings in chest X-ray images. 

To the best of our knowledge, the problem of generating image annotations for data curation purposes is among the less explored areas of medical image computing. Also, both architectures proposed here are novel combinations of network building blocks. One recent relevant work is the TieNet architecture proposed in \cite{DBLP:journals/corr/abs-1801-04334}. TieNet addressed the disease classification problem using joint information from image and text features during training and using image only for prediction. Another relevant work is \cite{7298952} which reports annotation from questions with an interactive process. We use reports both during training and prediction stages and the goal is mapping from text to image. Further, we model the anomaly localization as a regression problem on the polygonal bounding region coordinates. We have used these architectures on chest X-ray image annotation. We show that the availability of a large public dataset of X-ray images results in superior performance of the architecture with pre-trained image classifier.

\section{Methodology}
We describe two novel bimodal architectures for producing image level annotations. The ROI model selected here is a quadrilateral (a polygon with four edges). But the methodology can be generalized with more complex polygons. The outputs of both architectures are eight continuous values corresponding to the coordinates of vertices of the quadrilateral ROI. We describe the datasets used in this work, followed by the two architectures. 

\subsection{Datasets}
\textbf{DS1: } Dataset 1 is a public dataset from Indiana University Hospital that includes chest X-ray images from approximately 3500 unique patients \cite{demner2015preparingShort}. This collection is accompanied with corresponding reports. For a subset of this data, consisting of 494 images we performed image level annotations. We chose two groups of findings for image annotations. One was cardiomegaly which has a distinct signature showing an enlarged heart, and the second group was nodules, granuloma, and masses. These can appear in different areas of the lung and are more difficult to localize. Each of the 494 images paired with a text segment characterizing the anomaly, were presented to an experienced radiologist who marked a bounding quadrilateral around the finding. No limitation on size was enforced. The only limitation was to approximate the finding area with a four-corner convex shape. 

\textbf{DS2: } Dataset 2 is the public collection of chest X-ray images released by National Institutes of Health (NIH). Known as the Chest-Ray14 dataset \cite{DBLP:journals/corr/WangPLLBS17}, this comprises of 112,120 frontal-view X-rays. 51,708 images have one or more finding labels of the following 14: atelectasis, cardiomegaly, consolidation, edema, effusion, emphysema, fibrosis, hernia, infiltration, mass, nodule, pleural thickening, pneumonia, and pneumothorax. The remaining 60,412 images do not contain any of the 14 findings and are labeled as ``No Finding''. Several groups have recently reported classifiers trained using this dataset \cite{DBLP:journals/corr/HuangLW16a}. The ground truth labels are mined from clinical records and are not validated. The use of this dataset in our current work is for building a rich and large network that can act as the source of features. We do not have image level annotations or reports for this dataset. Therefore, we could not use this in our end to end architecture.   

\subsection{Architecture 1: Integrated CNN+LSTM, trained end to end}

\begin{figure}
\begin{centering}
\includegraphics[width = 120mm]{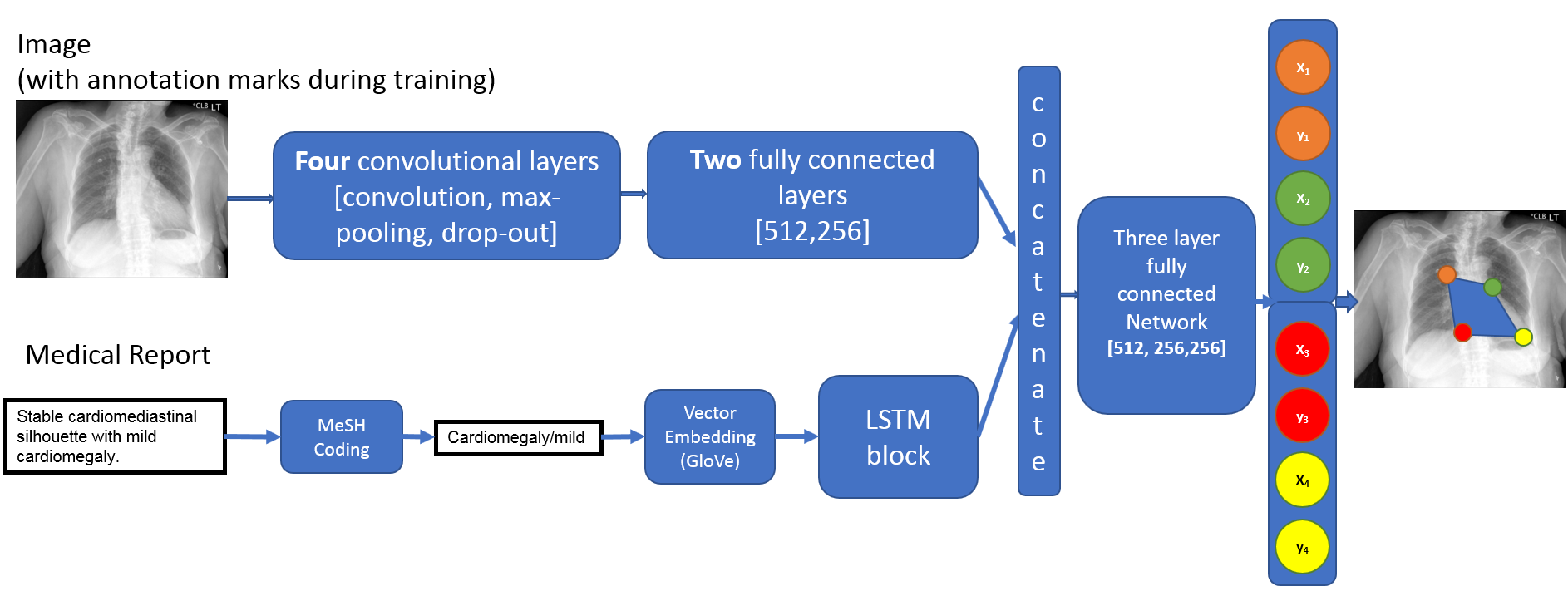}
\caption{Architecture 1: integrated text and imaging network.}
\label{fig:end2end}
\end{centering}
\end{figure}

Figure \ref{fig:end2end} displays our proposed integrated architecture. The image is fed through a convolutional neural network which learns relevant visual features. In the reported version, there are four convolutional layers, and two fully-connected layers. Dropout, batch normalization, and ReLU activation functions are utilized in the CNN. The final layer of the CNN is a 256 node layer that represents the dimensionality reduced version of the image to be merged with the text network output layer. 

On the text side, the findings and diagnoses in radiology reports have been manually coded with the standard Medical Subject Headings (MeSH) terms by domain experts \cite{demner2015preparingShort}. MeSH is the medical vocabulary controlled by the US National Library of Medicine. The next step is embedding of each MeSH term into a quantitative vector. For this purpose, we used the GloVe (Global Vectors for Word Representation) algorithm \cite{pennington2014glove}. Training of this embedding is performed on the term-term co-occurrence statistics from the 3500 reports in \textit{DS1}. The next block within the text pipeline is a long short-term memory (LSTM) layer with 128 units. This is followed by a 256 node fully connected layer together listed as the LSTM block. While the mesh terms and embedding models are frozen, this LSTM encoder is trained along with the CNN network with the objective of estimating the ROI coordinates.

In the final component of the architecture, the textual features from LSTM output and image features are merged by concatenation and fed through a multi-layer perceptron (MLP). The MLP is a 3 layer fully-connected network. The output is a series of [x,y] coordinates for the polygon vertices. The loss is calculated as the mean squared error between the predicted coordinates and the actual coordinates. 



\subsection{Architecture 2: Trained DenseNet121 + Doc2Vec}

Figure 2 demonstrates the components of the second architecture proposed in this paper. Given the limited size of the data with image level annotations, the CNN block of Architecture 1 is kept small and is still difficult to train. To solve this issue, we propose to learn the features characterizing the radiological findings within chest X-rays separately using \textit{DS2}. To this end, we trained a DenseNet architecture \cite{DBLP:journals/corr/HuangLW16a} with 121 layers. This is inspired by \cite{DBLP:journals/corr/abs-1711-05225}. However, in our architecture, the fully connected layer of DenseNet was replaced by a separate dense layer of dimension 1024 per finding class, followed by a sigmoid nonlinearity to create a 14-output network matching the 14 listed \textit{DS2} labels. To train this network, we randomly divided \textit{DS2} into a training set (80\%) and a validation set (20\%) and trained the network for 25 epochs with a batch size of 32, and with horizontal flip image augmentation. The performance was measured using the Area Under the Curve (AUC) by first generating the ROC curve per finding and averaging the AUC across all findings. 

This pre-trained network was used as a source of features in Architecture 2. The image features were extracted from the output of the global average pooling (GAP) layer that operates on the last convolution layer of the CheXNet architecture. The GAP layer produced a feature vector of 1024 elements.

On the text side, we also replaced the embedding and trainable LSTM block with a pre-trained paragraph vector (PV) model that supplies a 128 dimensional feature vector given a text segment. We used the unsupervised method known as \textit{PV-DBOW} to quantify text segments \cite{pmlr-v32-le14}. Unlike sequence learning through LSTM, PV-DBOW ignores context words in the input and predicts words randomly sampled from a text segment in the output. The model is capable of embedding input word sequences of variable length, and the learned representations are word-order independent, for which the encoding of a set of MeSH terms would be a good use case.  We trained \textit{PV-DBOW} using MeSH terms extracted from all 3500 reports available. Training was performed using stochastic gradient descent via backpropagation.

The two independently produced feature vectors for a text and image pair are then concatenated into a 1152 dimensional vector. After the input layer, there are three fully connected layers, with 1024, 512, and 128 neurons in each. Dropout is used throughout the network. The output layer feeds into the eight dimensional output representing the ROI coordinates. The network is trained using the mean squared error as measured by distance between estimated and annotated vertices. Backpropagation is used with Adam optimizer.

\begin{figure}
\begin{centering}
\includegraphics[width = 120 mm]{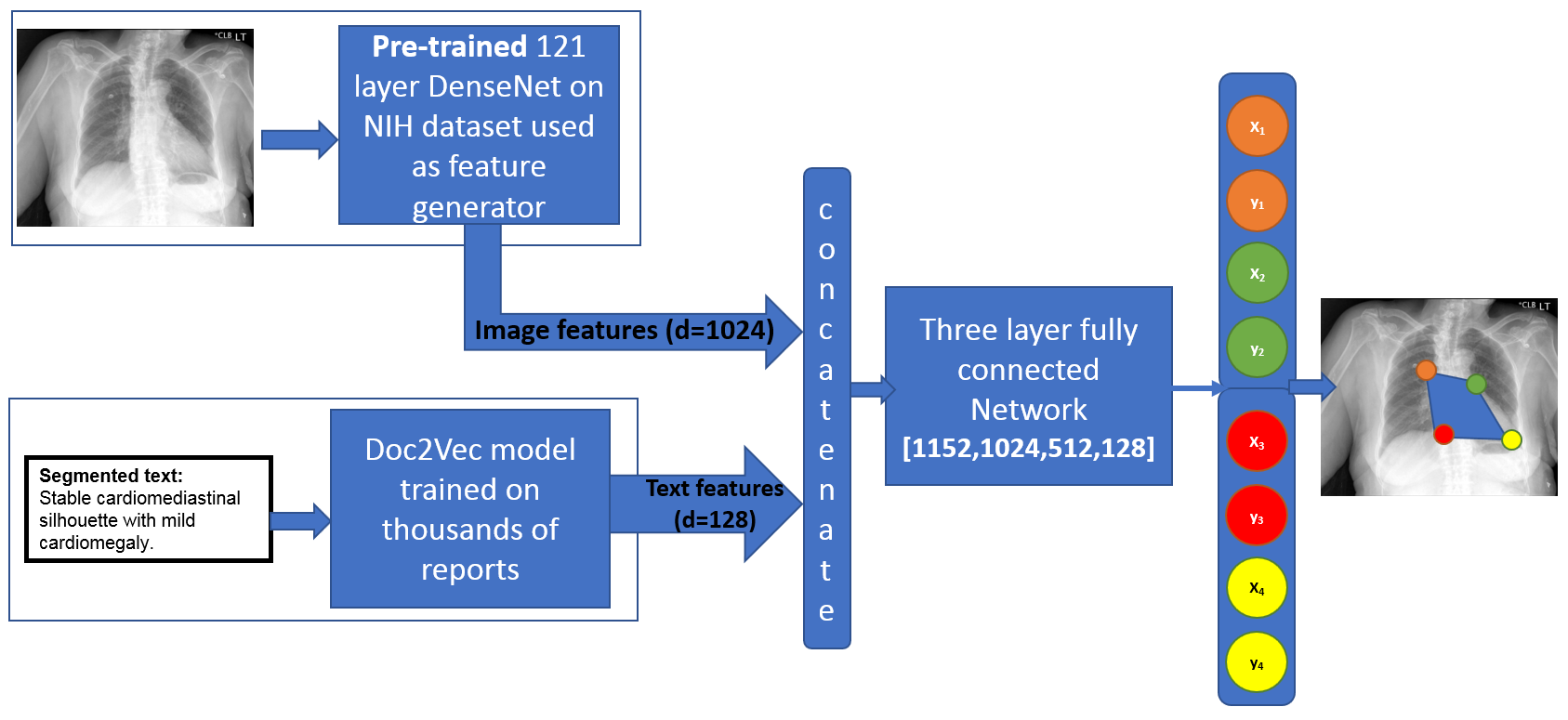}
\caption{Architecture 2: Image and text networks are trained separately. The resulting feature vectors are used to train a network that estimates ROI coordinates.}
\end{centering}
\end{figure}

\textbf{Experiments and performance metrics:} All images were re-sized to 256x256. We included an additional 800 images from \textit{DS1} that showed no clinically significant finding. For these images, the coordinates of the corresponding ROI were set to [0,0] for all four corners. These were used along with the 494 images with findings to train Architecture 1 and the final block of Architecture 2. One sanity check for the proposed solution is to not return a marking when there is no finding. For performance metrics, we report the centroid distance between the predicted and actual polygon normalized by image size. In experimenting with each of the two architectures, we performed 10 fold cross validation and report the average. All networks were trained on NVIDIA Titan X GPUs. 


\section{Results and Discussion}

\textbf{Architecture 1: }The average distance between the centroids of the annotated and estimated quadrilaterals obtained from the Architecture 1 network was 7.2$\pm$5.1\% of the image size (measured only for the images with finding). The cardiomegaly cases were correctly annotated. The network also correctly returned values within 1\% of [0,0] coordinates for all corners in all of the cases with a ``no finding'' label. For mass or nodules findings the estimated ROIs are close to ground truth, but not always overlapping. 



\textbf{Architecture 2: } For Architecture 2, it is important to first understand the performance of the DenseNet feature generator on predicting labels for its classification task. The measured average AUC of the model over the 14 labels in \textit{DS2} produced by our training process is 0.79. Table \ref{Table:ROCplaceholder} lists the AUCs for the 14 labels. The result is comparable to the those reported in \cite{DBLP:journals/corr/HuangLW16a} and demonstrates that the feature learning process for the findings, including cardiomegaly and mass, is fairly successful.

\begin{table}
  \centering
\caption{The area under ROC curves for the 14 labels present in the \textit{DS2} dataset obtained from a 121 layer DenseNet 
trained as the feature generator of Architecture 2.}
\label{Table:ROCplaceholder}

\begin{tabular}{|c|c|c|c|}
\hline
Atelectasis :\textbf{ 0.77}& Cardiomegaly:\textbf{ 0.89} &Effusion \textbf{0.86} & Infiltration: \textbf{0.70} \\
\hline
Nodule: \textbf{0.67} & Pneumonia: \textbf{0.73} &Pneumothorax: \textbf{0.81} & Consolidation: \textbf{0.80} \\
\hline

Emphysema: \textbf{0.84} & Fibrosis: \textbf{0.76} & Pleural thickening : \textbf{0.72} & Hernia: \textbf{0.78}  \\
\hline

Edema: \textbf{0.90} & Mass:\textbf{ 0.79} & \textbf{Average:} \textbf{0.79} &\\
\hline

\end{tabular}
\end{table}

\begin{figure}
\begin{centering}
\includegraphics[width = 100mm]{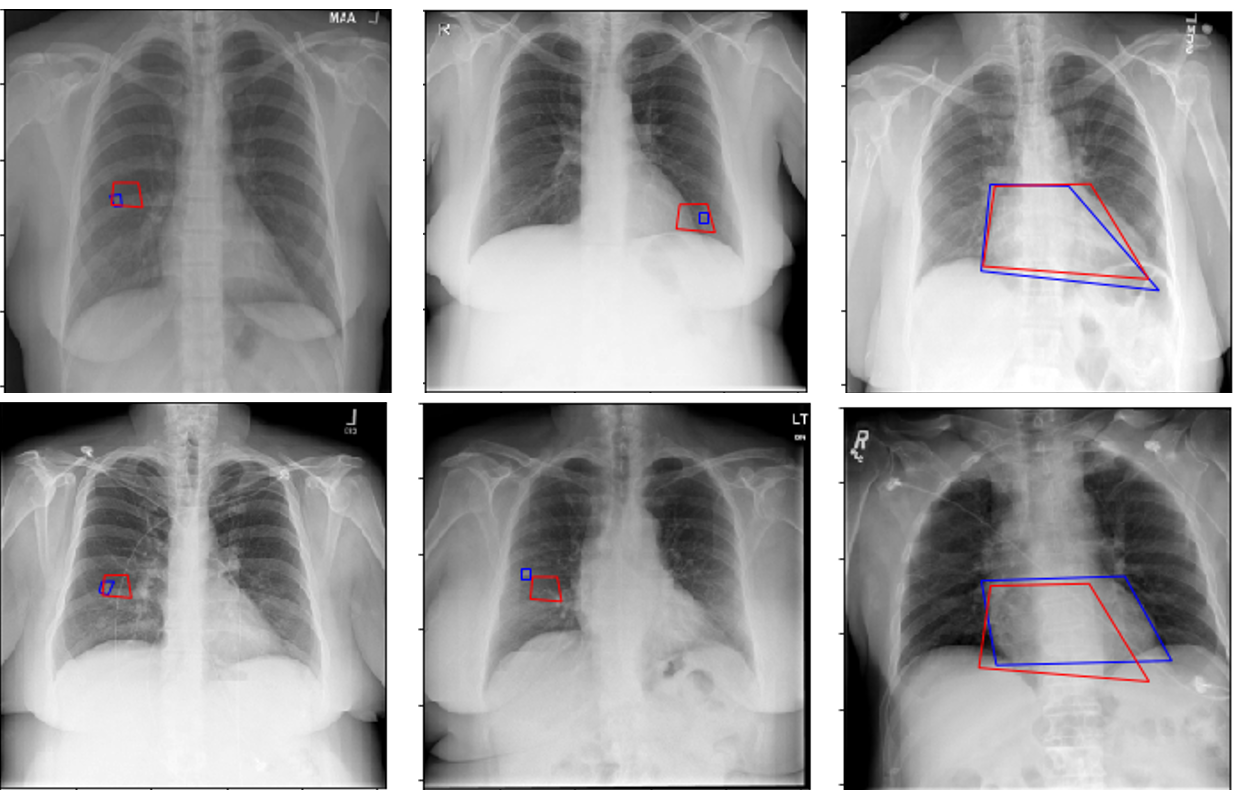}
\caption{Sample results from Architecture 2. Red is the estimated quadrilateral, blue is the one marked by a radiologist.}
  \label{fig:transferResultImages}
\end{centering}
\end{figure}

The fully trained Architecture 2 network returned an average distance between the centroids of the annotated and estimated quadrilaterals that was equivalent to 5.1$\pm$4.0\% of the image width. This is a significant improvement compared to Architecture 1. Examples in Figure \ref{fig:transferResultImages} show that in this case, despite the over-estimations of ROI size in cases of mass and nodule, the results are generally accurate and the estimated ROI includes the ground truth. The cardiomegaly cases (right column of Figure \ref{fig:transferResultImages}) show very accurate results. The overall Dice coefficient was 61\% compared to 46\% for Architecture 2. Note that given the approximate nature of the markings, the Dice coefficient is not a good performance measure here. It was notable that 100\% of the normal cases were mapped to the [0,0] corners and also all of the cardiomegaly cases were correctly mapped to the heart silhouette.

\textbf{Comparison with an image only network:} In both of the reported architectures, one can cut off the text side of the network and produce the ROI coordinates using only images. While this image only scenario was not our target, it is fair to ask if text features contribute at all to the outcome. When we used only the imaging arm of Architecture 1, the average centroid distance went from 7.2$\pm$5.1\% to 21$\pm$10.9\% and the results also clearly deteriorated visually. For Architecture 2, the elimination of text features resulted in an average centroid distance of 7.8$\pm$7.7\% compared to 5.1$\pm$4.0\% in the bimodal network. In both architectures, removing text features also results in a complete collapse of the specificity with over 88\% of normal images returning a non-zero ROI on Architecture 1, and 56\% on Architecture 2. The text features clearly contribute to the accuracy of ROI estimation in both architectures.

\section{Conclusion}
We proposed two architectures for mapping findings from clinical reports onto the relevant region of the corresponding image. One design trains a CNN and an LSTM jointly. A second design relies on separate imaging and text networks trained on very large datasets as sources of features that are combined to predict the ROI coordinates. This second architecture provides more accurate results that can be used for building annotated datasets with minimal editing. We showed that the text arms of these bimodal networks clearly contribute to the accuracy of the generated annotations. We used a simple model of the ROI to ease the segmentation process. A more complex shape model can be adopted and estimated similarly. Furthermore, one can replace the final fully connected layers of each of the proposed architectures with a segmentation network and obtain a mask with no shape constraints. This can also accommodate the cases with multiple image findings which are not modeled in the current work.

\end{document}